\documentclass[conference]{ieeeconf}  % Comment this line out if you need a4paper
\IEEEoverridecommandlockouts
\usepackage{cite}
\usepackage{amsmath,amssymb,amsfonts}
\usepackage{algorithmic}
\usepackage{graphics,graphicx,color}
\usepackage{textcomp}
\usepackage{xcolor}
\makeatletter
\let\NAT@parse\undefined
\makeatother
\usepackage{hyperref}  %hyperref still needs to be put at the end!
\usepackage{url}
\usepackage{subcaption}
\newcommand{\citep}[1]{\cite{#1}}
\usepackage{adjustbox}

\usepackage{pgfplots}
\usetikzlibrary{arrows.meta}

\pgfplotsset{compat=newest,
    width=.7\linewidth,
    height=.35\linewidth,
    scale only axis=true,
    max space between ticks=25pt,
    try min ticks=5,
    every axis/.style={
        axis y line=left,
        axis x line=bottom,
        axis line style={thick,->,>=latex, shorten >=-.4cm}
    },
    every axis plot/.append style={thick},
    tick style={black, thick},
    every axis plot/.style={line width=1.8pt},
}
\usepgfplotslibrary{groupplots}
\usepgfplotslibrary{dateplot}

\tikzset{
    c0/.append style={ourgreen},
    c1/.append style={ourorange},
    c2/.append style={ourred},
    c3/.append style={ourblue},
    c4/.append style={ourbrown},
    c5/.append style={ourlightblue},
}

\usepackage{amsmath,amsfonts,bm}

\def\figref#1{Fig.~\ref{#1}}
\def\Figref#1{Figure~\ref{#1}}
\def\twofigref#1#2{Figures \ref{#1} and \ref{#2}}

\def\secref#1{Sec.~\ref{#1}}
\def\eqref#1{Eq.~\ref{#1}}
\newcommand{\tabref}[1]{Table~\ref{#1}}

\def\1{\bm{1}}

\DeclareMathAlphabet{\mathsfit}{\encodingdefault}{\sfdefault}{m}{sl}
\SetMathAlphabet{\mathsfit}{bold}{\encodingdefault}{\sfdefault}{bx}{n}

\def\gC{{\mathcal{C}}}

\def\gF{{\mathcal{F}}}

\def\gH{{\mathcal{H}}}
\def\gI{{\mathcal{I}}}

\def\gM{{\mathcal{M}}}

\def\gO{{\mathcal{O}}}

\def\gS{{\mathcal{S}}}

\def\gZ{{\mathcal{Z}}}

\newcommand{\E}{\mathbb{E}}

\usepackage{amsmath,amsfonts,bm}
\usepackage{graphicx}
\usepackage{algorithm}
\usepackage{algorithmic}
\usepackage{xspace}
\usepackage{booktabs}       % professional-quality tables
\usepackage{stmaryrd}

\usepackage{xspace}
\makeatletter
\DeclareRobustCommand\onedot{\futurelet\@let@token\@onedot}
\def\@onedot{\ifx\@let@token.\else.\null\fi\xspace}

\makeatother

\newcommand{\solo}{Solo 8\xspace}
\newcommand{\method}{CASSI\xspace}
\newcommand{\id}{imitation discriminator\xspace}
\newcommand{\sd}{skill discriminator\xspace}
\newcommand{\oc}{oracle classifier\xspace}

\newcommand{\idisc}{\ensuremath{d_{\psi}}}
\newcommand{\sdisc}{\ensuremath{q_{\phi}}}
\newcommand{\oih}{\ensuremath{\mathbf{o}^\mathrm{I}}}
\newcommand{\osh}{\ensuremath{\mathbf{o}^\mathrm{S}}}
\newcommand{\och}{\ensuremath{\mathbf{o}^\mathrm{C}}} % on purpose the same as \osh
\newcommand{\sh}{\ensuremath{\mathbf{s}}}
\newcommand{\mapi}{\ensuremath{f^\mathrm{I}}}
\newcommand{\maps}{\ensuremath{f^\mathrm{S}}}

\newcommand{\taus}{\ensuremath{\tau^\mathrm{S}}}

\newcommand{\oI}{o^\mathrm{I}}
\newcommand{\gOI}{\gO^\mathrm{I}}
\newcommand{\oS}{o^\mathrm{S}}
\newcommand{\gOS}{\gO^\mathrm{S}}

\newcommand{\HI}{H^\mathrm{I}}
\newcommand{\HS}{H^\mathrm{S}}
\newcommand{\HC}{H^\mathrm{C}}

\newcommand{\rI}{r^\mathrm{I}}
\newcommand{\rS}{r^\mathrm{S}}
\newcommand{\rD}{r^\mathrm{D}}
\newcommand{\rT}{r^\mathrm{T}}
\newcommand{\rR}{r^\mathrm{R}}

\newcommand{\wI}{w^\mathrm{I}}
\newcommand{\wS}{w^\mathrm{S}}
\newcommand{\wD}{w^\mathrm{D}}
\newcommand{\wT}{w^\mathrm{T}}
\newcommand{\wR}{w^\mathrm{R}}

\newcommand{\wGP}{w^\mathrm{GP}}

\definecolor{ourblue}{rgb}{0.368,0.507,0.71}
\definecolor{ourorange}{rgb}{0.881,0.611,0.142}
\definecolor{ourgreen}{rgb}{0.56,0.692,0.195}
\definecolor{ourred}{rgb}{0.923,0.386,0.209}
\definecolor{ourviolet}{rgb}{0.528,0.471,0.701}
\definecolor{ourbrown}{rgb}{0.772,0.432,0.102}
\definecolor{ourlightblue}{rgb}{0.364,0.619,0.782}
\definecolor{ourdarkgreen}{rgb}{0.572,0.586,0.}

\newcommand{\ourlegendtable}{
{\small 
\begin{tabular}{l@{\qquad}l@{\qquad}l}
    {\color{ourgreen}\rule[.5ex]{2em}{1.5pt}} crawl &
    {\color{ourorange}\rule[.5ex]{2em}{1.5pt}} walk &
    {\color{ourred}\rule[.5ex]{2em}{1.5pt}} trot \\
    {\color{ourblue}\rule[.5ex]{2em}{1.5pt}} leap & 
    {\color{ourbrown}\rule[.5ex]{2em}{1.5pt}} wave & 
    {\color{ourlightblue}\rule[.5ex]{2em}{1.5pt}} stilt 
\end{tabular}
}
}

\title{\LARGE \bf
Versatile Skill Control via Self-supervised Adversarial Imitation of Unlabeled Mixed Motions}

\author{Chenhao Li$^{1}$, Sebastian Blaes$^{1}$, Pavel Kolev$^{1}$, Marin Vlastelica$^{1}$, Jonas Frey$^{2}$, and Georg Martius$^{1}$%
\thanks{$^{1}$Max Planck Institute for Intelligent Systems, Tübingen, Germany
{\tt\small \{firstname.lastname\}@tuebingen.mpg.de}}
\thanks{$^{2}$Robotic Systems Lab, ETH Zurich, Zurich, Switzerland
{\tt\small jonfrey@ethz.ch}}
}
\begin{document}

\maketitle
\thispagestyle{empty}
\pagestyle{empty}

\begin{abstract}
Learning diverse skills is one of the main challenges in robotics. To this end, imitation learning approaches have achieved impressive results. These methods require explicitly labeled datasets or assume consistent skill execution to enable learning and active control of individual behaviors, which limits their applicability. In this work, we propose a cooperative adversarial method for obtaining single versatile policies with controllable skill sets from unlabeled datasets containing diverse state transition patterns by maximizing their discriminability. Moreover, we show that by utilizing unsupervised skill discovery in the generative adversarial imitation learning framework, novel and useful skills emerge with successful task fulfillment. Finally, the obtained versatile policies are tested on an agile quadruped robot called \solo{} and present faithful replications of diverse skills encoded in the demonstrations.
\end{abstract}

% \begin{IEEEkeywords}
% Reinforcement learning, generative adversarial imitation learning, unsupervised skill discovery
% \end{IEEEkeywords}

\section{Introduction}
Reinforcement Learning (RL) has demonstrated its capability of learning diverse and complex skills for robotic platforms.
In the field of legged systems, RL has achieved success in learning-based quadrupedal locomotion control in challenging environments~\citep{hwangbo2019learning, lee2020learning, kumar2021rma, miki2022learning}.
Typically, deep RL techniques learn desired behaviors motivated by optimizing reward functions that are specifically tailored for the training task.
As a result, it can sometimes become very demanding to develop complex skills with reward engineering, where various terms of motivation and regularization need to be carefully designed and balanced.

Given the availability of some expert references, Imitation Learning (IL) allows an agent to learn to reproduce expert behavior.
In particular, Generative Adversarial Imitation Learning (GAIL,~\citep{ho2016generative}) employs Generative Adversarial Networks (GANs,~\citep{goodfellow2020generative}), which train a policy to deceive an \emph{imitation} discriminator that constantly tries to distinguish state transitions generated between the policy and the reference data distribution.
The output of the discriminator is used as a training signal that encourages the learning agent to generate similar behaviors to the demonstration~\citep{peng2021amp, li2022learning}.
However, given a large dataset of \emph{unlabeled} motion clips with diverse behaviors, extracting and learning \emph{individual} sensible skills can be challenging.
Such scenarios are commonly encountered in motion captures from creatures or underactuated experts, whose motion execution produces intrinsic noise and inconsistency and thus consists of miscellaneous behaviors.
Due to the unknown skill types in the dataset, supervised learning techniques fail to find correspondence between the policy and the individual skills it replicates.

\begin{figure}[t]
    \centering
    \includegraphics[width=1.0\linewidth]{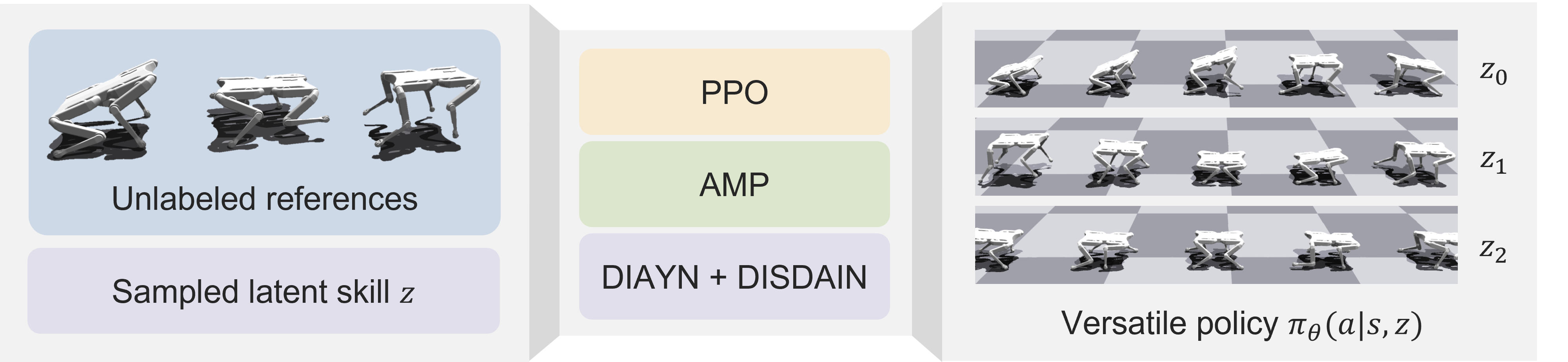}
    \caption{Method overview. Cooperative Adversarial Self-supervised Skill Imitation (\method) enables skill extraction and learning from unlabeled diverse references by motivating diversity in a generative adversarial imitation learning 
    % (\eg AMP) 
    setting using unsupervised learning techniques. 
    % (\eg DIAYN and DISDAIN). 
    The resulting versatile policy allows active control of individual skills.
    Supplementary videos and implementation details for this work are available at \url{https://sites.google.com/view/icra2023-cassi/home}.}
    \label{fig:method_overview}
\end{figure}

% For this reason, we propose to use unsupervised skill discovery methods to explore and learn individual skills simultaneously within the scope defined by the imitation dataset.
% However, for the purpose of acquiring knowledge that can be later applied to a variety of tasks, truly intelligent agents should be able to learn useful skills even without supervision.

% Unsupervised skill discovery methods in natural language processing~\citep{radford2019language, brown2020language} and computer vision~\citep{chen2020simple, chen2020big} have already obtained promising results in learning useful skills even without supervision. But its potential has been not yet fully realized in the learning of behavior, especially for real robotic applications.
In this case, unsupervised learning methods could help to address the problem.
Unsupervised skill discovery has enabled learning agents to obtain intrinsic behaviors without explicit task rewards.
One promising attempt for unsupervised skill discovery is based on maximizing the \emph{discriminability} of skills represented by latent variables on which a policy is conditioned~\citep{gregor2016variational, eysenbach2018diversity, hansen2019fast, sharma2019dynamics, strouse2021learning}.
Policy training signals are often constructed from variational approximations of the mutual information between latent variables and traversed state histories using a learned \emph{skill} discriminator.
Such mutual information encodes both the diversity in state transitions and the predictability of them given known latent skills~\citep{sharma2019dynamics}.
In this setting, strong discriminating performances are expected cooperatively by both the policy and the discriminator, where the latter is trained with supervised learning techniques.
The resulting policy is capable of producing consistently distinguishable behavioral patterns, or \emph{skills}.
% These skills can be evaluated zero-shot or serve as primitives in hierarchical RL setups to constitute high-level task fulfillment~\citep{eysenbach2018diversity, hansen2019fast, sharma2019dynamics}.
% To increase the robustness of learned behaviors to environmental perturbations, unsupervised skill learning objectives can also be combined with task rewards or regularizations, constraining the space of feasible policies to one that contains more safe and applicable skills~\citep{mahajan2019maven, kumar2020one}.

% As a result of lack of supervision, the skills developed solely based on discriminability maximization can be arbitrary and may suffer from poor execution.
% This results in inconsistent and unstable behaviors that are hardly useful in downstream applications, which is particularly crucial in real robotic settings.

% However, the degree to which such confinement is imposed highly depends on how these task rewards are defined.
% While a sparse reward allows some space for unsupervised skill learning to help address challenges in exploration~\citep{eysenbach2018diversity}, a dense reward may overconstrain the policy space such that unsupervised exploration is significantly hindered. 

In this work, we present Cooperative Adversarial Self-supervised Skill Imitation (\method), where we show that by applying unsupervised skill discovery in a generative adversarial imitation learning framework, a quadrupedal system manages to learn versatile policies that extract sensible high-level skills from a reference dataset containing diverse unlabeled motions (\figref{fig:method_overview}).
The learned policy enables active control of the individual skills embedded in the reference dataset, which manifests itself by its execution of consistently distinguishable behaviors.
The policy is also deployed and tested on the real robot without further adaptation.
To the best of our knowledge, this is the first time that individual skills are extracted and learned simultaneously from an unlabeled dataset motivated by unsupervised RL algorithms.
In summary, our contributions include: 
\textbf{(i)} A cooperative adversarial self-supervised skill imitation method for extracting and learning diverse skills from unlabeled references.
\textbf{(ii)} Fidelity and diversity evaluation of learned versatile policies using an \oc.
\textbf{(iii)} Experimental validation in simulation and on a quadruped robot.
% Supplementary videos and implementation details for this work are available at \url{https://sites.google.com/view/icra2023-cassi/home}.

\section{Related Work}

Recent development in robotics has created many potential applications that require intelligent systems to not only make decisions but also perform expected physical movements.
% However, due to imperfect parameter settings or algorithmic limitations, a learning agent may frequently fail to identify the desired behavior~\citep{pathak2019self, cai2020provably}.
While learning a task is often formulated as an optimization problem, it has become widely recognized that having prior knowledge provided by an expert can result in more effective and efficient learning than attempting to solve the problem from scratch~\citep{billard2008robot, argall2009survey}.
% General IL methods encourage the learning agent to replicate expert behaviors with an imitation target, which is often parameterized as a tracking objective that aims to reduce the pose error between the simulated character and target poses from a reference motion~\citep{lee2010data, liu2010sampling, liu2016guided, peng2018deepmimic}.
% The performance of these tracking-based methods often depends on carefully designed objective functions.
% While these methods typically work well for replicating single motion clips, the necessary synchronization and alignment between the agent and a specific reference motion require massive human efforts when applied to large and diverse motion
% datasets~\citep{peng2018deepmimic, peng2018sfv, lee2019scalable}.

Instead of relying on a handcrafted imitation objective, GAIL techniques train an adversarial discriminator to distinguish between behaviors generated by the agent and expert demonstrations~\citep{abbeel2004apprenticeship, ho2016generative}.
The effectiveness of imitation is thus measured by the performance of the discriminator, whose output encodes the similarity descriptions. 
To extend the applicability of GAIL to accommodate large datasets of unstructured motion clips, \citep{peng2021amp} employ an adversarial RL procedure that automatically selects the motion to perform in the reference, dynamically interpolating and generalizing from the dataset.
While this method has achieved success in learning adaptive motions, \emph{active} control over individual skill execution relies still on labeled motion datasets.
To this end, one may either collect disparate reference motions and learn individual skills separately~\citep{li2022learning}, or stack the training of multiple discriminators corresponding to each skill in parallel~\citep{vollenweider2022advanced}.
Both methods require \emph{supervision} from prior knowledge of motion labels in the datasets.
However, given a large dataset of diverse \emph{unlabeled} motion clips, enabling autonomous acquisitions of individual skills requires unsupervised training techniques.

Unsupervised RL is motivated by intrinsic skill development of intelligent agents that are believed to learn in the absence of supervision in order to acquire repurposable task-agnostic knowledge.
These skills can be quickly and efficiently utilized when specific tasks are later defined.
% The goal of such unsupervised training is to discover a set of skills that are useful for solving various downstream tasks.
Achieving this goal requires specifying a learning objective that ensures that each skill individually is distinct and that the skills collectively explore large parts of the state space~\citep{eysenbach2018diversity}.
% Recent work has drawn connections between RL and information theory~\citep{haarnoja2017reinforcement, haarnoja2018soft}.
In this setting, mutual information is commonly established as a notion of empowerment of an intrinsically motivated agent~\citep{mohamed2015variational}.
It is shown that a discriminability objective is equivalent to maximizing the mutual information between the skill and some aspect of the induced trajectory~\citep{florensa2017stochastic, hausman2018learning}.
\citep{gregor2016variational, baumli2021relative} try to maximize the mutual information between the skill and initial and final states.
\citep{eysenbach2018diversity, strouse2021learning} attempt to maximize the mutual information between the skill and states along the trajectory.
\citep{sharma2019dynamics, campos2020explore} propose to maximize the mutual information between the skill and the following state conditioned on the current state.
With the skill diversity motivated by mutual information maximization in various settings, diverse skills are acquired with minimal supervision.
Finally, unsupervised RL algorithms have been shown effective in conjunction with task rewards, which serve as sparse guidance for generating regularized and robust behaviors~\citep{mahajan2019maven, kumar2020one}.
% This provides the possibility to extract sensible skills within the policy space defined by the adversarial imitation problem. 

\section{Approach}

In this section, we describe our method, which enables learning from unlabeled demonstrations in a generative adversarial setting using unsupervised skill discovery techniques.
In the following context, we denote the discriminator distinguishing state transitions from the policy and reference in the generative adversarial imitation learning as the \emph{\id{}} ($\idisc$), and the discriminator distinguishing different skills yielded by the policy in the unsupervised skill discovery as the \emph{\sd{}} ($\sdisc$).

\subsection{Generative Adversarial Imitation}

Our generative adversarial imitation framework is built upon the Adversarial Motion Prior (AMP, \citep{peng2021amp}).
In this framework, the \id{} reflects the imitation performance and thus specifies the training signal for the policy. 
It is crucial to select an appropriate set of features that provide effective learning feedback.
We consider the imitation observation space $\gOI$.
The reference demonstrations are formulated as sequences of $\oI_t \in \gOI$, where the full state space $\gS$ of the underlying Markov Decision Process can be mapped to the imitation observation space $\gOI$ with a function $\mapi: \gS \to \gOI$.
We denote trajectory segments of length $\HI$ preceding time $t$ by $\oih_t=(\oI_{t-\HI+1},\dots,\oI_{t})$ for the reference observations and $\sh_t=(s_{t-\HI+1},\dots,s_{t})$ for the states induced by the policy.
For clarity, we omit the time index in the following. 
To simplify notation, we write $\mapi(\sh)$ to express that each state in $\sh$ is mapped to $\gOI$.   
In our experiments, we select linear and angular velocities of the robot base in the robot frame, measurement of the gravity vector in the robot frame, the base height, and joint angular position and velocity as the observation space $\gOI$.
As such, the discriminator's goal in this setup is to distinguish samples of the policy transition distribution $d^\pi$ from the reference motion distribution $d^\gM$.

\subsubsection{Imitation Discriminator Formulation}

% Due to saturation regions of the sigmoid cross-entropy loss function, the original GAN min-max loss (CEGAN) formulation has been shown to suffer from vanishing gradients, which slows down training~\citep{arjovsky2017towards}.
% The policy will be unable to obtain any information if the discriminator performs excessively well and becomes saturated because it is constantly penalized for being far from the demonstrations.
Similar to~\citep{peng2021amp}, we use the least-squares GAN (LSGAN) loss~\citep{mao2017least} for discriminator optimization.
Using $\HI$-step inputs and a gradient penalty, the discriminator objective is formulated as
\begin{equation}
    \begin{aligned}
    \E_{d^{\gM}} \left [ \big(\idisc(\oih) - 1 \big)^2 \right ] & + \mathbb{E}_{ d^{\pi}} \left [ \big(\idisc(\mapi(\sh)) + 1 \big)^2 \right ] \\
    & + \wGP \E_{d^{\gM}} \left [ \big \|\nabla_{\oih} \idisc \left (\oih \right) \big \|_2^2 \right ],
    \label{eqn:imitation_discriminator_loss}
    \end{aligned}
\end{equation}
where the last term denotes the penalty for non-zero gradients on samples from the dataset with weight $\wGP$ to stabilize training~\citep{peng2021amp}.

Intuitively, the LSGAN loss forces the discriminator to output $+1$ for samples from the reference motion and $-1$ for those from the policy.
In contrast, the policy is trained to deceive the discriminator by generating transition patterns similar to those present in the reference dataset.
The reward function for training the policy is then given by
\begin{equation}
    \rI = \mathrm{max}\left[0, 1-0.25 \left(\idisc (\mapi (\sh) ) - 1 \right)^2\right].
    \label{eqn:imitation_reward}
\end{equation}
The imitation reward $\rI$ provides a well-scaled output bounded between $0$ and $1$ which eases downstream policy learning.

% \subsubsection{Probabilistic Reference State Initialization}

% Generative adversarial imitation learning may suffer from poor exploration.
% Especially at the beginning of the training, when the reference motions seem far beyond the agent's capability, the discriminator is prone to become too strong such that the imitation reward gets clipped to zero and thus does not provide any information to train the policy.
% Recent attempts have proposed alternatives to the discriminator loss and imitation reward formulation to learn highly dynamic motions with partially available reference data~\citep{li2022learning}.
% However, when the reference motions contain rich and instructive state distribution, they can be used to direct the agent during training~\citep{peng2018deepmimic}.
% Reference state initialization (RSI) samples a random state from the reference motion at the beginning of each episode and utilizes it to initialize the agent's state.
% In our work, we apply RSI with a probability of $0.8$ to guide the agent during learning, while allowing exploration freedom of transitions from the initial state to the reference motions.

\subsection{Mutual Information Maximization}

Our unsupervised skill discovery framework is built upon Diversity Is All You Need (DIAYN, \citep{eysenbach2018diversity}) with the Discriminator Disagreement Intrinsic reward (DISDAIN, \citep{strouse2021learning}) for exploration motivation.

In the unsupervised RL setting, the agent seeks to develop a set of skills, indexed by a latent variable $z \in \gZ$ and represented by a policy $\pi_\theta(\cdot \mid s, z)$ parameterized by $\theta$. The latent skills $z \sim p_z$ are sampled at the beginning of each trajectory and then fixed over the episode. 
Therefore, each skill $z$ represents a temporally extended behavior within a state sequence of length $\HS$.
Similarly, we consider a function $\maps: \gS \to \gOS$ that maps the full state space $\gS$ to the skill observation space $\gOS$.
Thus, the transition in the extracted features can be represented by $\osh_t=(\oS_{t-\HS+1},\dots,\oS_{t}) \sim \taus(\pi_\theta(z))$, where $\taus$ denotes the state visitation distribution along the trajectory.
Again, we omit the time index in the following for clarity. 
Clearly, trajectories sampled from the policy conditioned
on a particular skill depend on the selection of $z$.
We highlight this dependence by writing $\osh_z$.

\subsubsection{Skill Discriminator Observation Space}

Since skills refer to consistently distinguishable behavioral motifs in the states of interest, how a skill distinguishes itself from other skills, or \emph{discriminability} between skills, depends crucially on the definition of the skill observation space $\gOS$.
In unsupervised RL literature, skills are often identified over only simple state transition patterns (e.g. different moving directions, speeds, or reached positions)~\cite{gregor2016variational, eysenbach2018diversity, sharma2019dynamics, strouse2021learning}, as it is typically hard to extract high-level skill descriptions without any supervision.
In our work, as the policy space is constrained by an imitation target, we are able to define a skill observation space $\gOS$ that specifies high-level distinction (e.g. trot, leap).
% As such, the skill observation space $\gOS$ used in our work consists of only linear and angular velocities $v$, $\omega$ of the robot base in the robot frame. 
% The idea can be illustrated in \figref{fig:unsupervised_rl}.

% \begin{figure}
%     \centering
%     \includegraphics[width=0.5\linewidth]{example-image-a}
%     \caption{Unsupervised RL within a constrained policy space. The corners of the policy space are defined by the (potentially unknown) motions in the reference dataset. The unsupervised skill discovery tries to find a set in the policy space that maximizes diversity.}
%     \label{fig:unsupervised_rl}
% \end{figure}

Note that the selection of states in $\gOS$ depends on users' interest.
% Ideally, it should contain enough information to allow the \sd{} to distinguish state transitions $\osh_z$ between different high-level skills.
% Meanwhile, it should also avoid containing excessive information to rule out the possibility of the \sd{} making different predictions over state transitions $\osh_z$ induced by the same skill (which might be caused by overfitting to noise).
% However, given an unknown reference dataset, one may not have prior knowledge of what motions are contained in the dataset.
If we are interested in obtaining different base motion patterns, including only the base information in $\gOS$ suffices.
If we also aim to recover transitions with different joint configurations (e.g. different gaits) in the reference dataset, including joint information will help with discrimination.
% As a result, some efforts are required to find an appropriate selection of features in $\gOS$.
% In \secref{app:sec:discriminator_observation}, we provide details of the selected features in $\gOS$ that are used in our work.
In addition, we assume that the skill space $\gZ$ is discrete and with cardinality $N_z$, although much of the discussion may extend to continuous skill space.

\subsubsection{Variational Approximation of Mutual Information}

A large and growing class of objectives for unsupervised skill discovery are derived from maximizing the mutual information between the latent skill $z$ and the resulting trajectories of extracted features $\osh_z$.
The formulation is expressed as
\begin{equation}
\begin{aligned}
    \gF(\theta) & = \gI(z; \osh_z)
    % & = \gH(\osh_z) - \gH(\osh_z \mid z) \\
    = \gH(z) - \gH(z \mid \osh_z) \\
    & = \E_{z \sim p_z, \osh_z \sim \taus(\pi_\theta(z))} \left [ \log p(z \mid \osh_z) - \log p_z(z)\right].
\end{aligned}
\end{equation}
%
% Note from the second line that maximizing this objective corresponds to maximizing the diversity of transitions produced in the environment, which is denoted by the entropy $\gH(\osh_z)$ while making $z$ informative about the induced trajectory by minimizing the entropy $\gH(\osh_z \mid z)$.
%
In practice, a variational approximation of the intractable conditional distribution $p(z \mid \osh_z)$ with a learned parametric model $\sdisc(z \mid \osh_z)$ is often applied to obtain a lower bound of $\gF(\theta)$~\citep{barber2003information}.
% Namely, one seeks to maximize the objective
% \begin{equation}
%     \tilde{\gF}(\theta)
%     = \E_{z\sim p_{z}, \osh_{z}\sim\taus(\pi_{\theta}(z))}\left[\log\sdisc(z\mid\osh_{z})-\log p_{z}(z)\right].
% \end{equation}
The model $\sdisc(z \mid \osh_z)$ is commonly referred to as a discriminator, as it is trained to discriminate between skills from state transitions.
In our work, it is termed as the \emph{\sd{}}, to distinguish itself from the \emph{\id{}} ($\idisc$) used in the generative adversarial imitation setting.

Optimizing the lower bound with respect to the policy parameters $\theta$ corresponds to training the policy with a skill reward
\begin{equation}
    \rS = \log \sdisc(z \mid \osh_z) - \log p_z(z).
    \label{eqn:skill_reward}
\end{equation}
To learn a full repertoire of skills, the skill prior is typically
fixed to be uniform~\citep{eysenbach2018diversity, achiam2018variational, baumli2021relative}, in which case $- \log p_z(z) = \log N_z$.
An arbitrary discriminator which simply ignores the trajectory will give a zero skill reward.
In contrast, a perfect discriminator will yield a skill reward of $\log N_z$~\citep{strouse2021learning}.

To tighten the lower bound, the discriminator should approximate the true distribution $p(z \mid \osh_z)$, which results in a supervised learning setting where the training data is collected from policy roll-outs.
In our work, the \sd{} is implemented as a classifier whose objective is to assign the correct skill to a given style observation by minimizing the cross-entropy classification loss. 
% In our work, the \sd{} is implemented as a classifier whose objective is defined as the cross-entropy loss of the classification performance.

As such, both the policy and the \sd{} expect a strong discriminating performance by maximizing and tightening the lower bound of the mutual information cooperatively.

\subsubsection{Policy Exploration Bonus}

Ideally, the policy presents different state transitions conditioned on different skills.
The discriminator receives these transitions and attempts to decode the original skill $z$.
However, at the beginning of the training, before the policy relates its behavior to the skill, the agent may yield similar transitions under different latent variables $z$.
As the skill is sampled randomly, this will generate mislabeled data which confuses the training of the discriminator.
When the discriminator fails to distinguish the transition patterns, the skill reward will be trivial and thus does not generate any information for the policy to diversify its behaviors.
This dead loop is termed as the \emph{pessimistic exploration} problem in unsupervised skill discovery~\citep{strouse2021learning}.
The reason behind this is that when the policy tries to maximize the skill reward $\rS$, it tends to reduce the \emph{epistemic} uncertainty by visiting the same states it has visited before.
To encourage exploration, \citep{strouse2021learning} propose the use of a discriminator ensemble of size $N$ and a resulting DISDAIN reward that compensates for the increase of the epistemic uncertainty when the policy visits new states: 
\begin{equation}
    \rD = \gH \left[ \dfrac{1}{N} \sum_{i=1}^N q_{\phi_i} (z \mid \osh_z) \right ] - \dfrac{1}{N} \sum_{i=1}^N \gH \left[ q_{\phi_i} (z \mid \osh_z) \right ],
    \label{eqn:disdain_reward}
\end{equation}
where $\phi_i$ corresponds to the parameters of the ensemble member $i$.

For trajectories with rich training data for the skill discriminator, the ensemble members should agree and result in a zero DISDAIN reward.
In contrast, for state transitions of high discriminator disagreement, this reward will be positive and thus encourage exploration.

\subsection{Overview}

Note that the imitation reward $\rI$, the skill reward $\rS$, and the DISDAIN reward $\rD$ are task-agnostic.
This allows the training of primitive skills that facilitate downstream learning such as hierarchical RL when a task is later specified.
Alternatively, we are able to learn high-level tasks alongside self-supervised skill extraction. 
In this setting, we define task reward $\rT$, in addition to some regularization terms $\rR$ that enforce stable policy outputs on the real platform.
Putting everything together, the total reward that the policy receives encompasses five parts
\begin{equation}
    r = \wT \rT + \wI \rI + \wS \rS + \wD \rD + \wR \rR,
    \label{eqn:total_reward}
\end{equation}
where $w$ denotes the weight of each term and $\wD = \wR = 1.0$ is kept constant throughout our work.

The joint optimization of the policy and the \id{} utilizes \emph{adversarial} training.
The \id{} tries to distinguish state transitions between the agent and the reference motion, whereas the policy tries to make this difficult by generating similar behaviors.
Meanwhile, the joint training of the policy and the \sd{} forms a \emph{cooperative} game.
The agent samples a skill and tries to present a distinct behavior for the ease of discrimination by the \sd{}, and the \sd{} tries to decode the original skill and yield training signals to reward the policy for diversifying its skills.

\begin{figure}
    \centering
    \includegraphics[width=1.0\linewidth]{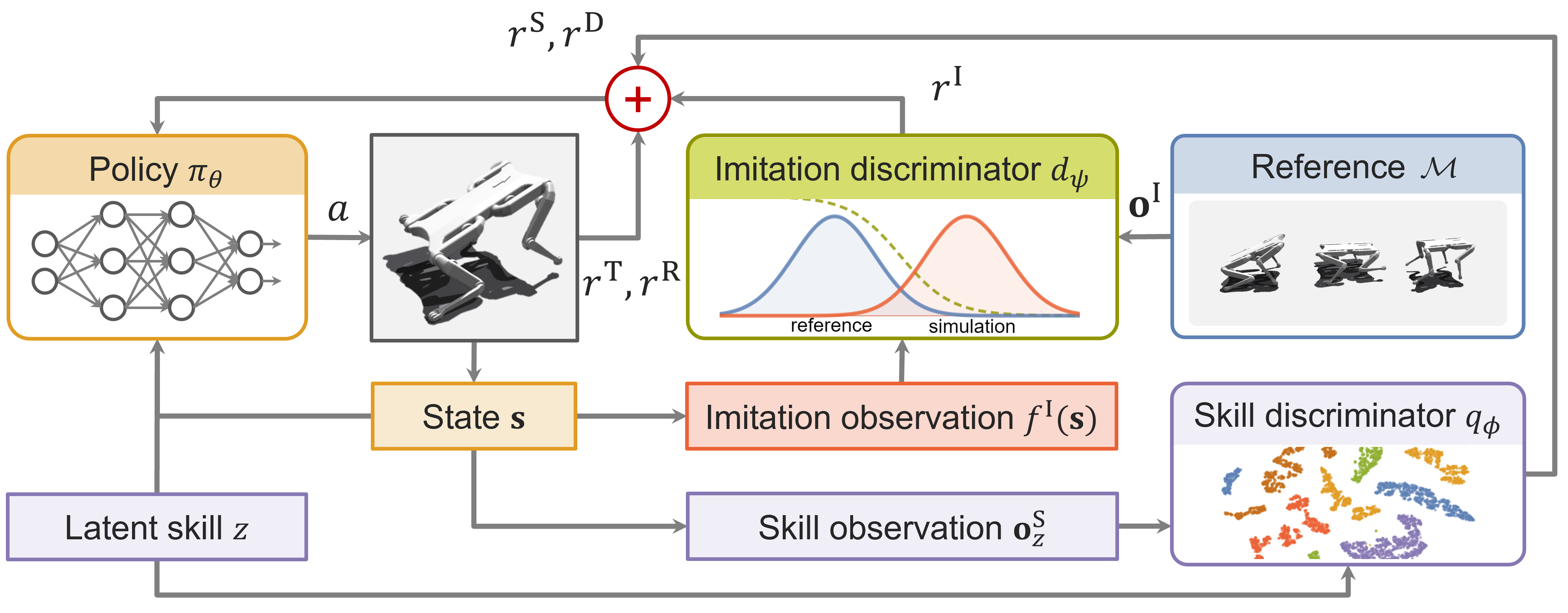}
    \caption{System overview. Given an unlabeled dataset, an imitation discriminator learns sampled state transition patterns from demonstrations. At the beginning of each episode, a latent skill variable $z$ is sampled as input to the policy.
    % motivates the policy to generate distinct motions. 
    A skill discriminator is trained to decode the original skill from these motions. The policy is rewarded for making the skills distinguishable.
    %, whose discriminating performance is promoted to encourage discriminability among skill executions.
    }
    \label{fig:system_overview}
\end{figure}

\Figref{fig:system_overview} provides a schematic overview of our method, and an algorithm overview is detailed in Algorithm \ref{algorithm1}.

\begin{algorithm}[htbp]
    \caption{\method}
    \footnotesize
    \label{algorithm1}
    \begin{algorithmic}[1]
        \STATE \textbf{Input}: unlabeled reference dataset $\mathcal{M}$; imitation and skill feature maps $\mapi, \maps$; transition horizons $\HI, \HS$; latent skill cardinality $N_z$
        \STATE initialize state transition buffer $\tau$, replay buffer $B$
        \FOR{learning iterations $ = 1,2,\dots$}
                \STATE sample latent skill variable $z \sim p_z$
                \STATE collect $\sh$ with policy $\pi_\theta$ conditioned on skill $z$
                \STATE extract features $\osh_z$ by applying $\maps$ to $\sh$
                \STATE compute $\rI, \rS, \rD$ according to Equations \ref{eqn:imitation_reward}, \ref{eqn:skill_reward} and \ref{eqn:disdain_reward}
                \STATE calculate total reward $r$ according to \eqref{eqn:total_reward}
                \STATE fill replay buffer $B$ with $ \left (\sh, z,\osh_z \right)$
            \FOR{policy learning epoch $ = 1,2,\dots$}
                \STATE sample transition mini-batches $b^\pi\sim B$
                \STATE update $V$ and $\pi_\theta$ with PPO or another RL algorithm
            \ENDFOR
            \FOR{\id{} learning epoch $ = 1,2,\dots$} 
                \STATE sample transition mini-batches $b^\pi\sim B$ and $b^\gM \sim \gM$ 
                \STATE update $\idisc$ using $b^\pi$ and $b^\mathcal{M}$ according to \eqref{eqn:imitation_discriminator_loss}
            \ENDFOR
            \FOR{\sd{} learning epoch $ = 1,2,\dots$} 
                \STATE sample transition mini-batches $b^\pi\sim B$ with bagging
                \STATE update $\sdisc$ using $b^\pi$ with cross-entropy or loss
            \ENDFOR
        \ENDFOR
    \end{algorithmic}
\end{algorithm}

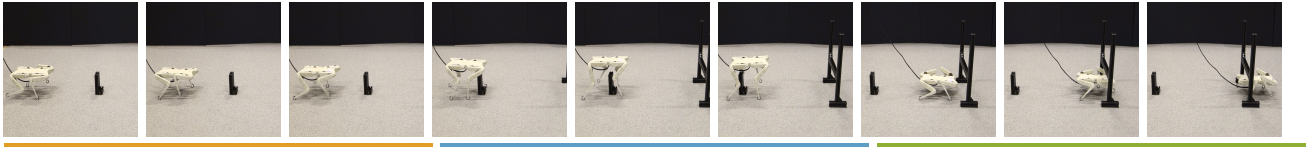
\begin{figure*}
    {
    \centering
    \hspace{1em}
    \includegraphics[width=.1\linewidth,height=.1\linewidth]{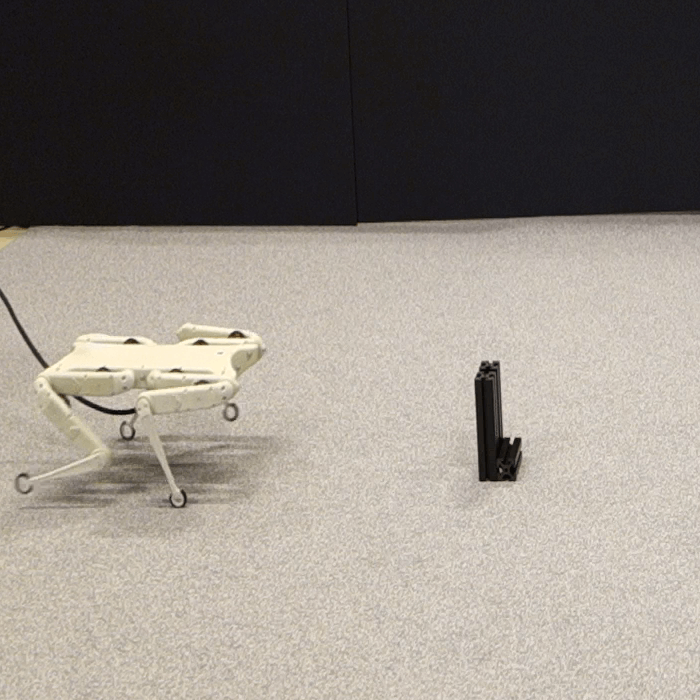}
    \includegraphics[width=.1\linewidth,height=.1\linewidth]{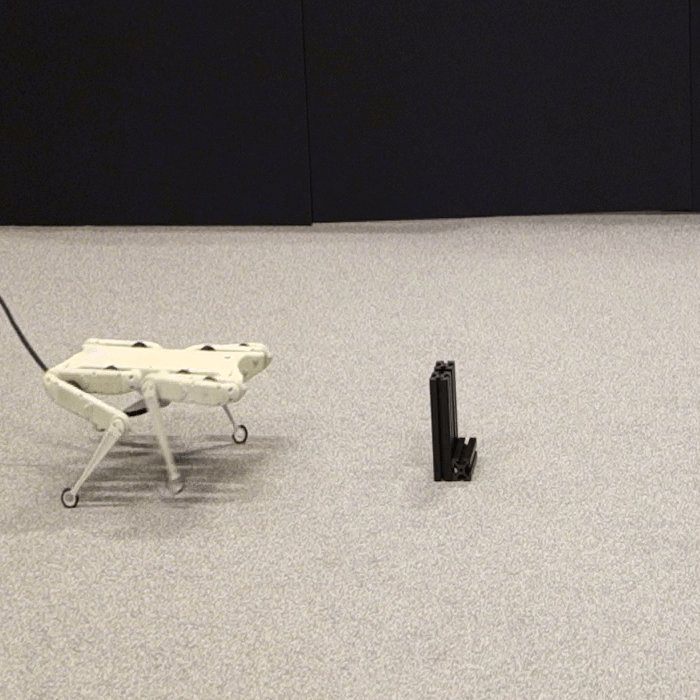}
    \includegraphics[width=.1\linewidth,height=.1\linewidth]{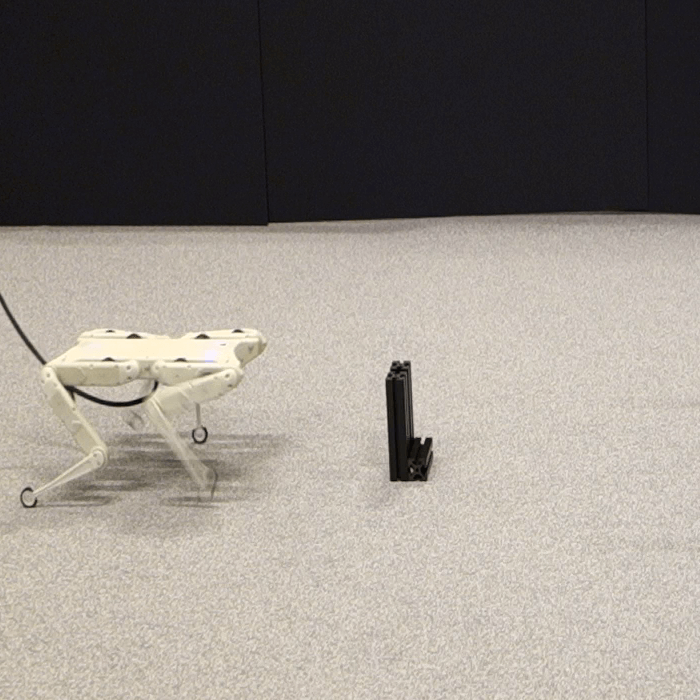}
    \includegraphics[width=.1\linewidth,height=.1\linewidth]{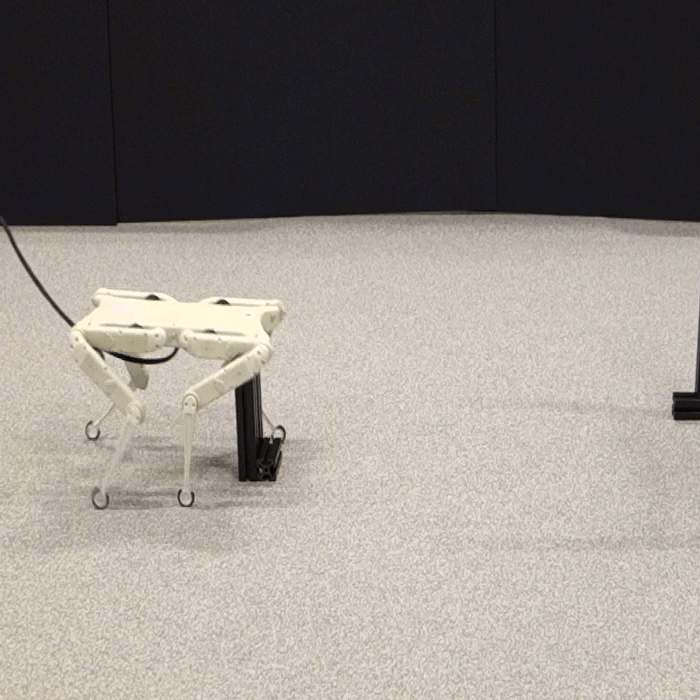}
    \includegraphics[width=.1\linewidth,height=.1\linewidth]{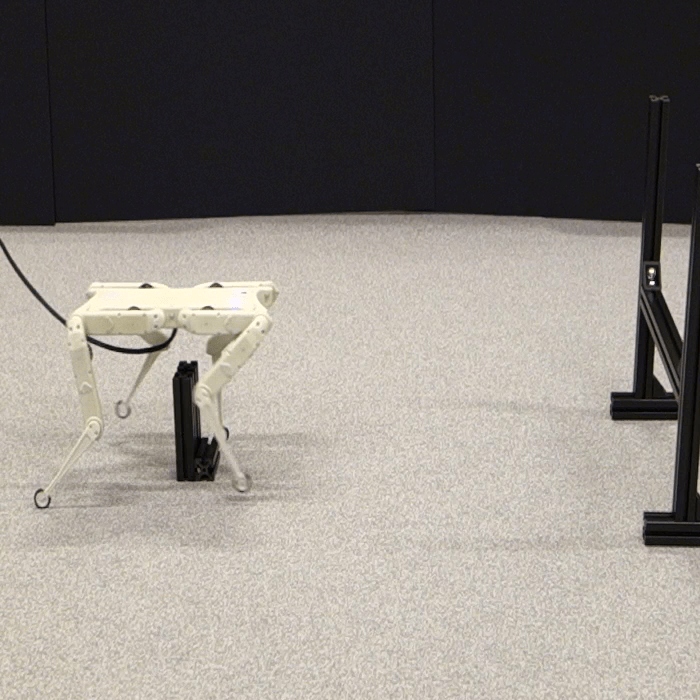}
    \includegraphics[width=.1\linewidth,height=.1\linewidth]{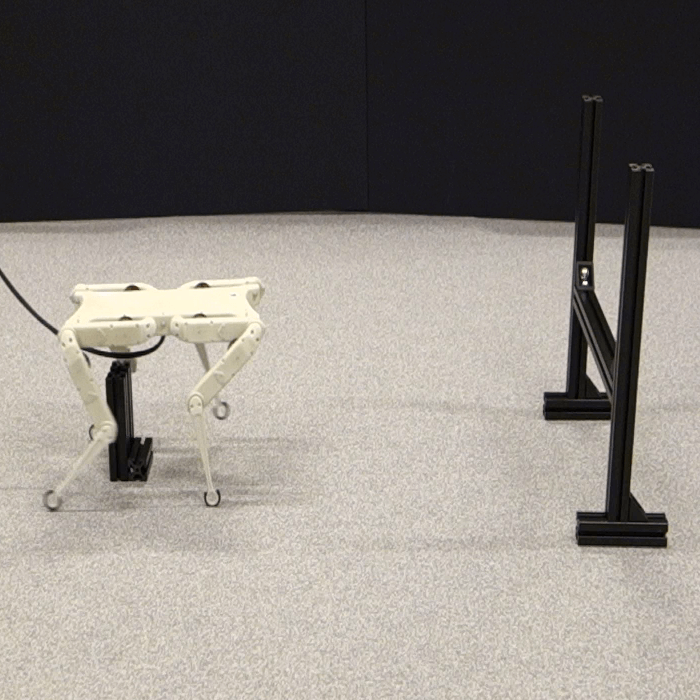}
    \includegraphics[width=.1\linewidth,height=.1\linewidth]{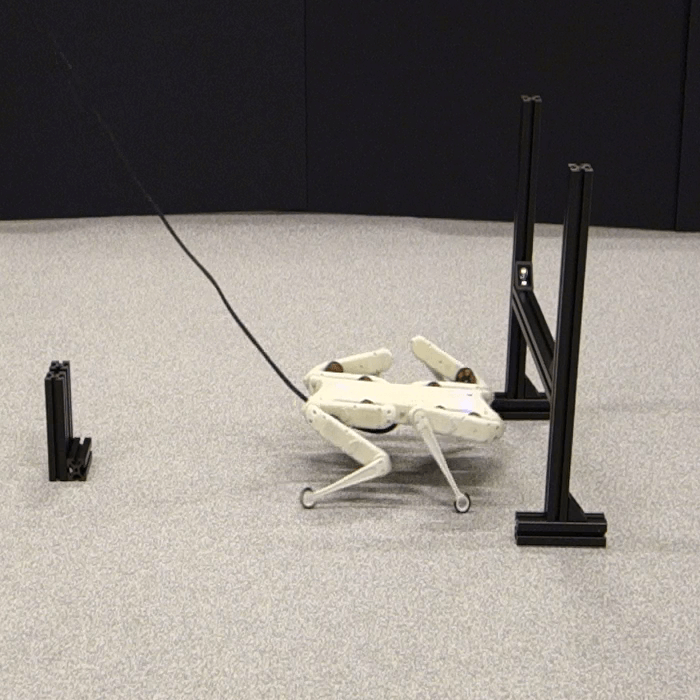}
    \includegraphics[width=.1\linewidth,height=.1\linewidth]{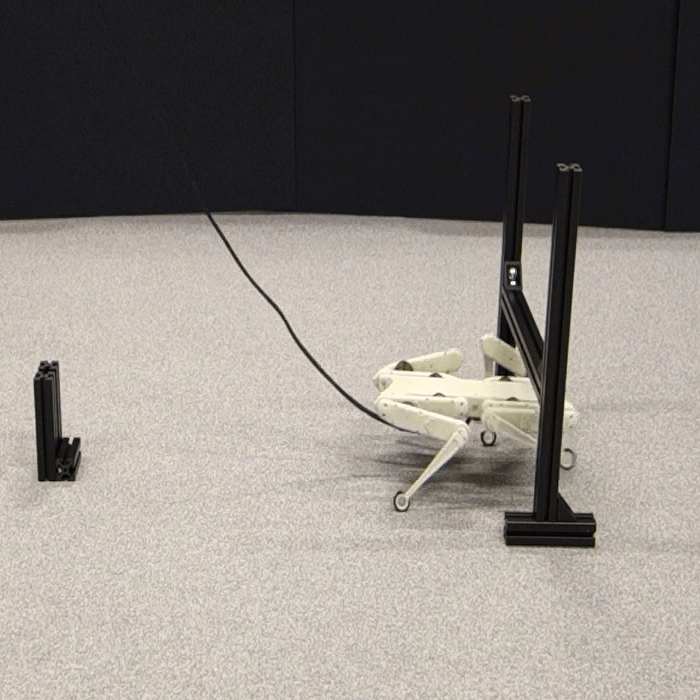}
    \includegraphics[width=.1\linewidth,height=.1\linewidth]{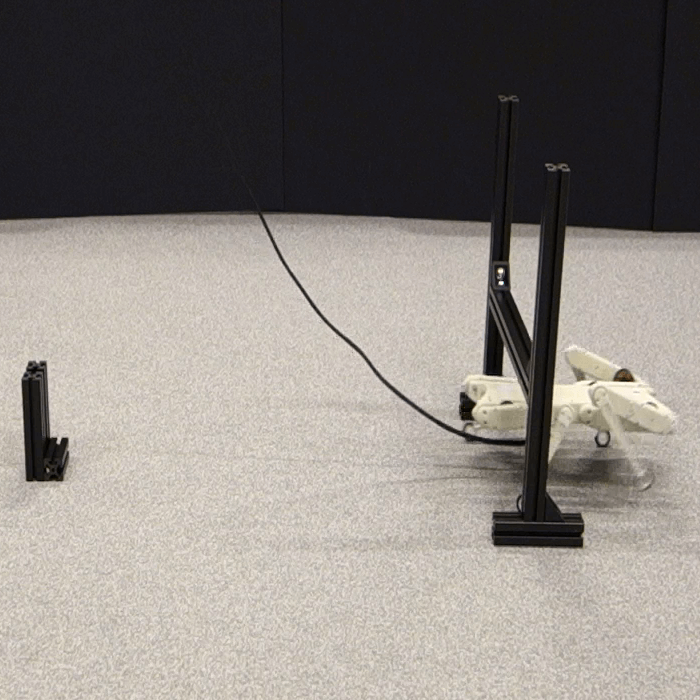}
    }
    
    \vspace{-.8em}
    \hspace{1em}
    \begin{tikzpicture}
    \draw[fill=ourorange,draw=none] (0,0) rectangle ++(5.7,0.05);
    \draw[fill=ourlightblue,draw=none] (5.8,0) rectangle ++(5.7,0.05);
    \draw[fill=ourgreen,draw=none] (11.61,0) rectangle ++(5.7,0.05);
    \end{tikzpicture}
    \caption{Motion sequence composed of multiple skills, executed on the real \solo{} in an obstacle avoidance environment.}
    \label{fig:motion_sequence}
\end{figure*}

\section{Experiments}

We evaluate our method on the \solo{} robot, an open-source research quadruped robot that performs a wide range of physical actions~\citep{grimminger2020open}, in simulation and on the real system.

% \begin{figure}
%     \centering
%     \begin{subfigure}[t]{.24\linewidth}
%         \includegraphics[width=\linewidth]{figs/solo8.png}
%     \end{subfigure}
%     \begin{subfigure}[t]{.71\linewidth}
%         \includegraphics[width=\linewidth]{figs/isaac_gym.png}
%     \end{subfigure}
% 	\caption{\solo{} (left). Wave motion in Isaac Gym (right).}
%     \label{fig:solo8}
% \end{figure}

The unlabeled motion dataset is constructed using mixed trajectories induced by individual expert policies learned in a previous work~\citep{li2022learning}, including {\bf\color{ourgreen}{crawl}}, {\bf\color{ourorange}{walk}}, {\bf\color{ourred}{trot}}, {\bf\color{ourblue}{leap}}, {\bf\color{ourbrown}{wave}}, and {\bf\color{ourlightblue}{stilt}}. These colors are used consistently throughout the paper.
For each motion, 1000 trajectories are recorded in simulation from parallel robot instances with randomized mechanical properties.
Each trajectory contains 120 consecutive time steps.
To break down the consistency in skill execution, these consecutive state transitions are sliced into small motion clips of horizon length~8.
% An example motion sequence is shown in \figref{fig:motion_sequence}.
In all of our experiments, we use Proximal Policy Optimization (PPO,~\citep{schulman2017proximal}) in Isaac Gym~\citep{makoviychuk2021isaac}.

\subsection{Skill Extraction and Imitation}

In this section, we set task reward weight $\rT = 0$ and compare \method with spectral clustering and AMP in extracting and learning skills from unlabeled datasets.

\subsubsection{Spectral Clustering}

A potential strategy to obtain individual skills from unlabeled references is to perform pre-clustering on the dataset and learn each separately.
To test its effectiveness, we perform spectral clustering with $k$-Nearest Neighbors on the reference dataset, compute the error rate with respect to the true labels, and compare it with the \sd learned during the training of our method.
We report the best result in \tabref{table:spectral_clustering}.

\begin{table}[b]
    \centering
    \caption{Learned \sd vs. spectral clustering.}
    \begin{tabular}{l|c|cccccc}
    \toprule
        & \method & \multicolumn{6}{c}{Spectral Clustering}\\
        \midrule
        Horizon & $8$ & $8$ & \color{gray}$20$ & \color{gray}$30$ & \color{gray}$40$ & \color{gray}$60$ & \color{gray}$120$ \\
        \midrule
        Error \% & $\textbf{0.016}$ & $\textbf{70.2}$ & \color{gray}$29.1$ & \color{gray}$27.3$ & \color{gray}$21.1$ & \color{gray}$17.9$ & \color{gray}$ 0.08$ \\
    \bottomrule
    \end{tabular}
    \label{table:spectral_clustering}
\end{table}

\begin{figure}[t]
    \centering
    \includegraphics[width=.8\linewidth]{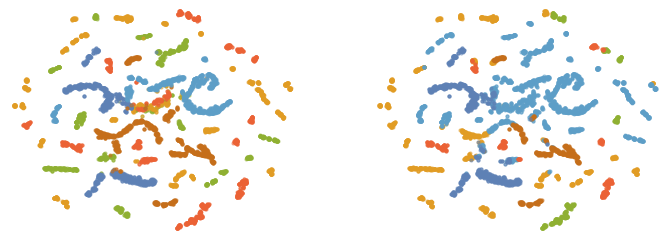}\\
    \vspace{0.2cm}
    \ourlegendtable
    \caption{Predicted motion clusters by spectral clustering (right) with a horizon of 8 on a random subset of the dataset (1/10 of tuples from each motion) compared with ground truth labels (left). The prediction error of spectral clustering reports $43.57\%$ on this subset.}
\label{fig:clustering_subtrajectories}
\end{figure}

The result reveals that the \sd{} learned using our method with a horizon of only 8 achieves better discriminating performance than spectral clustering on the whole trajectories.
This indicates that for spectral clustering to work with decent accuracy, strong assumptions of consistent skill execution over trajectories with much longer horizons must be made.
The spectral clustering label assignment is visualized using t-SNE in \figref{fig:clustering_subtrajectories} next to the ground-truth labels.

\subsubsection{Oracle Classifier}

As we construct the dataset by mixing individual expert motion clips, we are able to evaluate the diversity and fidelity of the learned skills by training an \oc $C$ using ground-truth labels.
We denote the true label by $c \in \gC$ and the true number of skills present in the dataset by $N_c$, where $\gC$ denotes the true skill space. 
In our experiments, $\gC = \{{\color{ourgreen}{\text{crawl}}}, {\color{ourorange}{\text{walk}}}, {\color{ourred}{\text{trot}}}, {\color{ourblue}{\text{leap}}}, {\color{ourbrown}{\text{wave}}}, {\color{ourlightblue}{\text{stilt}}}\}$ and $N_c = 6$.
Note that $N_c$ is unknown to the design of the \sd.
Thus, the predefined number of skills $N_z$ is a tuning parameter.
%that does not necessarily align with $N_c$.

We write $p ( c \mid \och_z)$ to represent the probability of the \oc predicting the true skill $c$ given a state trajectory of horizon length $\HC$, which is induced by the policy conditioned on a sampled latent skill $z$.
When the policy conditioned on a skill $z$ is able to generate a state trajectory that aligns well with the transition patterns exerted by a certain skill $c$ in the true skill space $\gC$, the probability $p ( c \mid \och_z )$ is large for this specific skill $c \in \gC$ and small for all other skills $c' \in \gC\backslash\{c\}$.
We thus can quantify the diversity and fidelity of the skills learned by the policy $\pi_\theta$, using the following entropy expressions:
\begin{equation}
    \text{div}(\pi_\theta) = \gH \left[ \dfrac{1}{N_z} \sum_{z=1}^{N_z}  p ( c \mid \och_z ) \right],\label{eq:diversity_metric}
\end{equation}
\begin{equation}
    \text{fid}(\pi_\theta) = - \dfrac{1}{N_z} \sum_{z=1}^{N_z} \gH \left[ p ( c \mid \och_z ) \right]\label{eq:fidelity_metric}.
\end{equation}

Intuitively, fidelity is maximized if the policy is very certain that every generated trajectory corresponds to a true skill present in the reference dataset when conditioned on a latent skill.
In contrast, diversity is maximized if the policy is able to generate trajectories corresponding to different true skills present in the reference dataset when conditioned on different latent skills.

\begin{figure}[b]
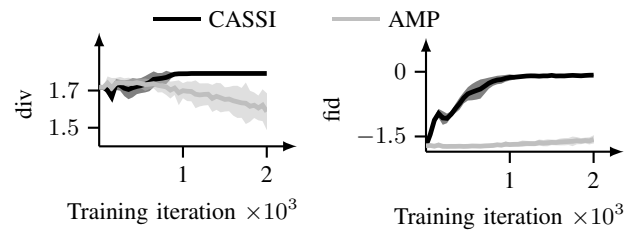

    \centering
    {\small {{\color{black} \rule[.5ex]{2em}{1.5pt}} \method  \qquad
{\color{gray!50} \rule[.5ex]{2em}{1.5pt}} AMP}}
\vspace{1.1em}

\pgfplotsset{
    width=0.6\linewidth,
    height=0.3\linewidth,
    xtick={1000, 2000},
    xticklabels={1, 2},
}
\small
    \begin{subfigure}[]{.43\linewidth}
       \input{plots/sgl/diversity/diversity}%
    \end{subfigure}
    \hspace{0.5em}
    \begin{subfigure}[]{.43\linewidth}
       \input{plots/sgl/predictibility/predictibility}%
    \end{subfigure}
    \caption{Diversity (left, \eqref{eq:diversity_metric}) and fidelity (right, \eqref{eq:fidelity_metric}) of the learned skills for \method and AMP ($\wS=0$) over iterations.}
    \label{fig:div_fid}
\end{figure}

We visualize the training performance in terms of $\text{div}(\pi_\theta)$, $ \text{fid}(\pi_\theta)$, and $p ( c \mid \och_z )$ in \twofigref{fig:div_fid}{fig:iteration_curves} for $\wI=1.0$, $\wS=0.5$, and $N_z=6$.
Note that under the assumption $N_z = N_c$, the learned policy establishes a bijective correspondence $g: \gZ \to \gC$, where all reference motions are assigned to distinct latent variables.
In addition, the policy encounters skill collapse with AMP only, due to the absence of the skill reward $\rS$ which promotes discriminability among learned behaviors.
We further analyze cases with $N_z \neq N_c$ in \secref{sec:task_execution}.

\begin{figure}
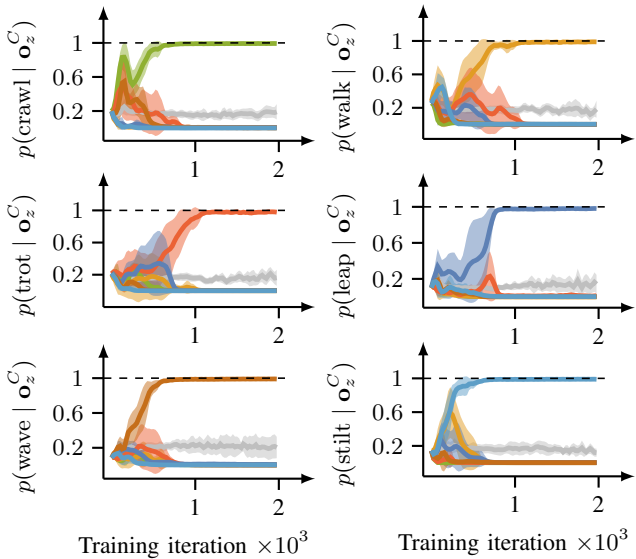

    \centering
%     {\small {{\color{ourgreen} \rule[.5ex]{2em}{1.5pt}} $z=0$  \qquad
% {\color{ourorange} \rule[.5ex]{2em}{1.5pt}} $z=1$ \qquad
% {\color{ourred}\rule[.5ex]{2em}{1.5pt}} $z=2$ \qquad
% {\color{ourblue} \rule[.5ex]{2em}{1.5pt}} $z=3$ \qquad
% {\color{ourbrown} \rule[.5ex]{2em}{1.5pt}} $z=4$ \qquad
% {\color{ourlightblue} \rule[.5ex]{2em}{1.5pt}} $z=5$}}
\vspace{0.7em}
\pgfplotsset{
    width=.65\linewidth,
    height=.375\linewidth,
    xmax=2000,xmin=0,
    ytick={0.2, 0.6, 1.0},
    xtick={1000, 2000},
    xticklabels={1, 2},
}
\tikzset{
    z0plt/.append style={ourgreen},
    z1plt/.append style={ourorange},
    z2plt/.append style={ourred},
    z3plt/.append style={ourblue},
    z4plt/.append style={ourbrown},
    z5plt/.append style={ourlightblue},
    grayplt/.append style={gray!50},
    z0fill/.append style={fill=ourgreen, fill opacity=0.5},
    z1fill/.append style={fill=ourorange, fill opacity=0.5},
    z2fill/.append style={fill=ourred, fill opacity=0.5},
    z3fill/.append style={fill=ourblue, fill opacity=0.5},
    z4fill/.append style={fill=ourbrown, fill opacity=0.5},
    z5fill/.append style={fill=ourlightblue, fill opacity=0.5},
    grayfill/.append style={fill=gray!50, fill opacity=0.5},
}
\small
\hspace{-2em}
\begin{subfigure}[]{0.43\linewidth}
    \input{plots/sgl/master_cls_over_time_aligned/master_cls_over_time_aligned_c_0}%
\end{subfigure}
\hspace{1em}
\begin{subfigure}[]{0.43\linewidth}
    \input{plots/sgl/master_cls_over_time_aligned/master_cls_over_time_aligned_c_1}%
\end{subfigure}
  
\hspace{-2em}
\begin{subfigure}[]{0.43\linewidth}
    \input{plots/sgl/master_cls_over_time_aligned/master_cls_over_time_aligned_c_2}%
\end{subfigure}
\hspace{1em}
\begin{subfigure}[]{0.43\linewidth}
    \input{plots/sgl/master_cls_over_time_aligned/master_cls_over_time_aligned_c_3}%
\end{subfigure}
  
\hspace{-2em}
\begin{subfigure}[]{0.43\linewidth}
    \input{plots/sgl/master_cls_over_time_aligned/master_cls_over_time_aligned_c_4}%
\end{subfigure}
\hspace{1em}
\begin{subfigure}[]{0.43\linewidth}
    \input{plots/sgl/master_cls_over_time_aligned/master_cls_over_time_aligned_c_5}%
\end{subfigure}
    \caption{Training performance for \method (colored) and AMP (grey, $\wS=0$) over iterations. AMP witnesses a skill collapse where the policy generates single state transition patterns regardless of the latent skill $z$. 
    In contrast, \method encourages each latent skill $z$ to converge to a distinct reference motion $c$.
    % , and each reference motion is assigned to a distinct skill. 
    % See \figref{fig:div_fid} for a comparison between the two methods w.r.t. the diversity and fidelity of the learned skills. 
    As the mapping between $\gZ$ and $\gC$ may differ between runs%
    % due to the unsupervised learning nature
    , the plots show results after aligning correspondence across seeds.}
    \label{fig:iteration_curves}
\end{figure}

% In practice, motion-captured data may contain various transitions with different proportions.
% We investigate the robustness of \method being able to still recover certain motions by changing their relative proportions within the dataset.
% To this end, we vary the number of trajectories of the wave motion used in the reference dataset while keeping those of other motions unchanged.

% \begin{figure}
% \pgfplotsset{
%     width=.6\linewidth,
%     height=.25\linewidth,
%     }
% \small
%     \centering
%     \input{plots/sgl/ratio/ratio}
%     % \begin{subfigure}[]{.49\linewidth}
%     %     \input{plots/sgl/ratio/ratio_ratio_10}%
%     % \end{subfigure}
%     % \begin{subfigure}[]{.49\linewidth}
%     %     \input{plots/sgl/ratio/ratio_ratio_1000}%
%     % \end{subfigure}
%     % \includegraphics[width=0.5\linewidth]{example-image-a}
%     \caption{Single histogram of p(wave$|$z) after 10 (left) and 1000 (right) trajectories.}
%     \label{fig:varying_ratio}
% \end{figure}

\subsection{Task Execution} \label{sec:task_execution}

In this section, we evaluate the skills learned using \method with an additional velocity tracking task, which is achieved by maximizing the reward 
$\rT = \exp\{ -\sigma^{-2}(u-v_x)^2 \}$,
% \begin{equation}
%     \rT = \exp\left\{ -\dfrac{(u-v_x)^2}{\sigma^2} \right\},
% \end{equation}
where $u$ denotes the velocity command sampled within the range $[0, 1]$.
We denote by $v_x$ the linear velocity of the robot base, expressed in the robot frame, and set the constant scaling factor $\sigma^2 = 0.25$.
The task reward weight is set constant $\wT = 1.0$.

\subsubsection{Novel Skill Discovery} \label{sec:novel_skill_discovery}

In practice, motion types presented in the reference dataset are often unknown.
Without prior knowledge of the actual number of component motions $N_c$, the choice of the latent skill cardinality $N_z$ may require additional effort.
We evaluate skill extraction performance for different $N_z$ using the \oc.

\begin{figure}[b]
\pgfplotsset{
    width=.68\linewidth,
    height=.35\linewidth,
}
\tikzset{
    c0/.append style={fill=ourgreen},
    c1/.append style={fill=ourorange},
    c2/.append style={fill=ourred},
    c3/.append style={fill=ourblue},
    c4/.append style={fill=ourbrown},
    c5/.append style={fill=ourlightblue},
}
\small
    \centering
    \captionsetup[subfigure]{oneside,margin={-1cm,0cm}}
    \begin{subfigure}[t]{.32\linewidth}
        % This file was created with tikzplotlib v0.10.1.
\begin{tikzpicture}[trim axis left]

\begin{axis}[
tick align=outside,
tick pos=left,
x grid style={darkgray176},
xmin=-0.59, xmax=3.59,
xtick style={color=black},
y grid style={darkgray176},
ymin=0, ymax=1.0499999947437,
ytick style={color=black},
xtick={0,1,2,3},
ylabel={\(\displaystyle p(c \mid \mathbf{o}^{C}_{z})\)},
xlabel={\(\displaystyle z\)},
ytick={0.2, 0.6, 1.0},
y label style={at={(-0.3,0.5)}},
]
\draw[draw=none, c0] (axis cs:-0.4,0) rectangle (axis cs:0.4,0.9319761372);
\draw[draw=none, c1] (axis cs:-0.4,0.9319761372) rectangle (axis cs:0.4,0.9810192706);
\draw[draw=none, c2] (axis cs:-0.4,0.9810192706) rectangle (axis cs:0.4,0.984834221571);
\draw[draw=none, c3] (axis cs:-0.4,0.984834221571) rectangle (axis cs:0.4,0.996671550749);
\draw[draw=none, c4] (axis cs:-0.4,0.996671550749) rectangle (axis cs:0.4,0.999361595252);
\draw[draw=none, c5] (axis cs:-0.4,0.999361595252) rectangle (axis cs:0.4,0.9999999854495);
\draw[draw=none, c0] (axis cs:0.6,0) rectangle (axis cs:1.4,0.003227740293);
\draw[draw=none, c1] (axis cs:0.6,0.003227740293) rectangle (axis cs:1.4,0.015938403193);
\draw[draw=none, c2] (axis cs:0.6,0.015938403193) rectangle (axis cs:1.4,0.02309945608);
\draw[draw=none, c3] (axis cs:0.6,0.02309945608) rectangle (axis cs:1.4,0.12627585216);
\draw[draw=none, c4] (axis cs:0.6,0.12627585216) rectangle (axis cs:1.4,0.99494132966);
\draw[draw=none, c5] (axis cs:0.6,0.99494132966) rectangle (axis cs:1.4,0.999999994994);
\draw[draw=none, c0] (axis cs:1.6,0) rectangle (axis cs:2.4,0.002995036164);
\draw[draw=none, c1] (axis cs:1.6,0.002995036164) rectangle (axis cs:2.4,0.606789677364);
\draw[draw=none, c2] (axis cs:1.6,0.606789677364) rectangle (axis cs:2.4,0.990563091684);
\draw[draw=none, c3] (axis cs:1.6,0.990563091684) rectangle (axis cs:2.4,0.995816845999);
\draw[draw=none, c4] (axis cs:1.6,0.995816845999) rectangle (axis cs:2.4,0.998551206611);
\draw[draw=none, c5] (axis cs:1.6,0.998551206611) rectangle (axis cs:2.4,0.9999999862179);
\draw[draw=none, c0] (axis cs:2.6,0) rectangle (axis cs:3.4,0.002406721664);
\draw[draw=none, c1] (axis cs:2.6,0.002406721664) rectangle (axis cs:3.4,0.011982625681);
\draw[draw=none, c2] (axis cs:2.6,0.011982625681) rectangle (axis cs:3.4,0.016601658252);
\draw[draw=none, c3] (axis cs:2.6,0.016601658252) rectangle (axis cs:3.4,0.432899531052);
\draw[draw=none, c4] (axis cs:2.6,0.432899531052) rectangle (axis cs:3.4,0.436253479151);
\draw[draw=none, c5] (axis cs:2.6,0.436253479151) rectangle (axis cs:3.4,0.999999970851);
\end{axis}

\end{tikzpicture}%
    \caption{$N_z = 4$}
    \label{fig:varying_Nz_4}
    \end{subfigure}
    \hspace{-1.5em}
    \begin{subfigure}[t]{.32\linewidth}
        % This file was created with tikzplotlib v0.10.1.
\begin{tikzpicture}

\begin{axis}[
tick align=outside,
tick pos=left,
x grid style={darkgray176},
xlabel={\(\displaystyle z\)},
xmin=-0.69, xmax=5.69,
xtick style={color=black},
y grid style={darkgray176},
ymin=0, ymax=1.04999999832199,
ytick style={color=black},
xtick={0,1,2,3,4,5},
ytick={0.2, 0.6, 1.0},
ymajorticks=false,
]
\draw[draw=none, c0] (axis cs:-0.4,0) rectangle (axis cs:0.4,0.9704690689);
\draw[draw=none, c1] (axis cs:-0.4,0.9704690689) rectangle (axis cs:0.4,0.99136734149);
\draw[draw=none, c2] (axis cs:-0.4,0.99136734149) rectangle (axis cs:0.4,0.9923930983233);
\draw[draw=none, c3] (axis cs:-0.4,0.9923930983233) rectangle (axis cs:0.4,0.9973294282653);
\draw[draw=none, c4] (axis cs:-0.4,0.9973294282653) rectangle (axis cs:0.4,0.9994850675511);
\draw[draw=none, c5] (axis cs:-0.4,0.9994850675511) rectangle (axis cs:0.4,0.9999999984019);
\draw[draw=none, c0] (axis cs:0.6,0) rectangle (axis cs:1.4,0.0046121131);
\draw[draw=none, c1] (axis cs:0.6,0.0046121131) rectangle (axis cs:1.4,0.9916549963);
\draw[draw=none, c2] (axis cs:0.6,0.9916549963) rectangle (axis cs:1.4,0.99405301422);
\draw[draw=none, c3] (axis cs:0.6,0.99405301422) rectangle (axis cs:1.4,0.996137658887);
\draw[draw=none, c4] (axis cs:0.6,0.996137658887) rectangle (axis cs:1.4,0.998826813167);
\draw[draw=none, c5] (axis cs:0.6,0.998826813167) rectangle (axis cs:1.4,0.9999999668413);
\draw[draw=none, c0] (axis cs:1.6,0) rectangle (axis cs:2.4,0.00208941355);
\draw[draw=none, c1] (axis cs:1.6,0.00208941355) rectangle (axis cs:2.4,0.02422784881);
\draw[draw=none, c2] (axis cs:1.6,0.02422784881) rectangle (axis cs:2.4,0.98975620141);
\draw[draw=none, c3] (axis cs:1.6,0.98975620141) rectangle (axis cs:2.4,0.997250704276);
\draw[draw=none, c4] (axis cs:1.6,0.997250704276) rectangle (axis cs:2.4,0.999196397048);
\draw[draw=none, c5] (axis cs:1.6,0.999196397048) rectangle (axis cs:2.4,0.9999999847575);
\draw[draw=none, c0] (axis cs:2.6,0) rectangle (axis cs:3.4,0.002781069703);
\draw[draw=none, c1] (axis cs:2.6,0.002781069703) rectangle (axis cs:3.4,0.004622056689);
\draw[draw=none, c2] (axis cs:2.6,0.004622056689) rectangle (axis cs:3.4,0.008157165631);
\draw[draw=none, c3] (axis cs:2.6,0.008157165631) rectangle (axis cs:3.4,0.989817025431);
\draw[draw=none, c4] (axis cs:2.6,0.989817025431) rectangle (axis cs:3.4,0.997974217678);
\draw[draw=none, c5] (axis cs:2.6,0.997974217678) rectangle (axis cs:3.4,0.999999976015);
\draw[draw=none, c0] (axis cs:3.6,0) rectangle (axis cs:4.4,0.0012266419341);
\draw[draw=none, c1] (axis cs:3.6,0.0012266419341) rectangle (axis cs:4.4,0.0044130777081);
\draw[draw=none, c2] (axis cs:3.6,0.0044130777081) rectangle (axis cs:4.4,0.0061498611313);
\draw[draw=none, c3] (axis cs:3.6,0.0061498611313) rectangle (axis cs:4.4,0.0075149765801);
\draw[draw=none, c4] (axis cs:3.6,0.0075149765801) rectangle (axis cs:4.4,0.9991176763801);
\draw[draw=none, c5] (axis cs:3.6,0.9991176763801) rectangle (axis cs:4.4,0.9999999928851);
\draw[draw=none, c0] (axis cs:4.6,0) rectangle (axis cs:5.4,0.0006640100705);
\draw[draw=none, c1] (axis cs:4.6,0.0006640100705) rectangle (axis cs:5.4,0.0032103312085);
\draw[draw=none, c2] (axis cs:4.6,0.0032103312085) rectangle (axis cs:5.4,0.004335330367);
\draw[draw=none, c3] (axis cs:4.6,0.004335330367) rectangle (axis cs:5.4,0.010330468867);
\draw[draw=none, c4] (axis cs:4.6,0.010330468867) rectangle (axis cs:5.4,0.015869103326);
\draw[draw=none, c5] (axis cs:4.6,0.015869103326) rectangle (axis cs:5.4,0.999999995526);
\end{axis}

\end{tikzpicture}%
    \caption{$N_z = 6$}
    \label{fig:varying_Nz_6}
    \end{subfigure}
    \hspace{-1.5em}
    \begin{subfigure}[t]{.32\linewidth}
        % This file was created with tikzplotlib v0.10.1.
\begin{tikzpicture}

\begin{axis}[
tick align=outside,
tick pos=left,
x grid style={darkgray176},
xlabel={\(\displaystyle z\)},
xmin=-0.79, xmax=7.79,
xtick style={color=black},
y grid style={darkgray176},
ymin=0, ymax=1.05000000893561,
ytick style={color=black},
xtick={0,1,2,3,4,5,6,7},
ytick={0.2, 0.6, 1.0},
ymajorticks=false,
]
\draw[draw=none, c0] (axis cs:-0.4,0) rectangle (axis cs:0.4,0.9911788468);
\draw[draw=none, c1] (axis cs:-0.4,0.9911788468) rectangle (axis cs:0.4,0.994119216812);
\draw[draw=none, c2] (axis cs:-0.4,0.994119216812) rectangle (axis cs:0.4,0.9949448898004);
\draw[draw=none, c3] (axis cs:-0.4,0.9949448898004) rectangle (axis cs:0.4,0.9971478731744);
\draw[draw=none, c4] (axis cs:-0.4,0.9971478731744) rectangle (axis cs:0.4,0.9996630140154);
\draw[draw=none, c5] (axis cs:-0.4,0.9996630140154) rectangle (axis cs:0.4,0.9999999945113);
\draw[draw=none, c0] (axis cs:0.6,0) rectangle (axis cs:1.4,0.003870445296);
\draw[draw=none, c1] (axis cs:0.6,0.003870445296) rectangle (axis cs:1.4,0.987109249896);
\draw[draw=none, c2] (axis cs:0.6,0.987109249896) rectangle (axis cs:1.4,0.992626635673);
\draw[draw=none, c3] (axis cs:0.6,0.992626635673) rectangle (axis cs:1.4,0.995493734586);
\draw[draw=none, c4] (axis cs:0.6,0.995493734586) rectangle (axis cs:1.4,0.998457285754);
\draw[draw=none, c5] (axis cs:0.6,0.998457285754) rectangle (axis cs:1.4,0.9999999735783);
\draw[draw=none, c0] (axis cs:1.6,0) rectangle (axis cs:2.4,0.0012881381693);
\draw[draw=none, c1] (axis cs:1.6,0.0012881381693) rectangle (axis cs:2.4,0.0061593382443);
\draw[draw=none, c2] (axis cs:1.6,0.0061593382443) rectangle (axis cs:2.4,0.9908021684443);
\draw[draw=none, c3] (axis cs:1.6,0.9908021684443) rectangle (axis cs:2.4,0.9974851934063);
\draw[draw=none, c4] (axis cs:1.6,0.9974851934063) rectangle (axis cs:2.4,0.9994644539133);
\draw[draw=none, c5] (axis cs:1.6,0.9994644539133) rectangle (axis cs:2.4,1.0000000085101);
\draw[draw=none, c0] (axis cs:2.6,0) rectangle (axis cs:3.4,0.002558816263);
\draw[draw=none, c1] (axis cs:2.6,0.002558816263) rectangle (axis cs:3.4,0.022881644438);
\draw[draw=none, c2] (axis cs:2.6,0.022881644438) rectangle (axis cs:3.4,0.034634301391);
\draw[draw=none, c3] (axis cs:2.6,0.034634301391) rectangle (axis cs:3.4,0.982462266791);
\draw[draw=none, c4] (axis cs:2.6,0.982462266791) rectangle (axis cs:3.4,0.989355021232);
\draw[draw=none, c5] (axis cs:2.6,0.989355021232) rectangle (axis cs:3.4,0.999999969726);
\draw[draw=none, c0] (axis cs:3.6,0) rectangle (axis cs:4.4,0.0013245474201);
\draw[draw=none, c1] (axis cs:3.6,0.0013245474201) rectangle (axis cs:4.4,0.0098462121771);
\draw[draw=none, c2] (axis cs:3.6,0.0098462121771) rectangle (axis cs:4.4,0.0136526176091);
\draw[draw=none, c3] (axis cs:3.6,0.0136526176091) rectangle (axis cs:4.4,0.0149961344302);
\draw[draw=none, c4] (axis cs:3.6,0.0149961344302) rectangle (axis cs:4.4,0.9989718662302);
\draw[draw=none, c5] (axis cs:3.6,0.9989718662302) rectangle (axis cs:4.4,0.9999999615929);
\draw[draw=none, c0] (axis cs:4.6,0) rectangle (axis cs:5.4,0.0005055769671);
\draw[draw=none, c1] (axis cs:4.6,0.0005055769671) rectangle (axis cs:5.4,0.0018617701759);
\draw[draw=none, c2] (axis cs:4.6,0.0018617701759) rectangle (axis cs:5.4,0.0026038986268);
\draw[draw=none, c3] (axis cs:4.6,0.0026038986268) rectangle (axis cs:5.4,0.0082435558618);
\draw[draw=none, c4] (axis cs:4.6,0.0082435558618) rectangle (axis cs:5.4,0.0104972393768);
\draw[draw=none, c5] (axis cs:4.6,0.0104972393768) rectangle (axis cs:5.4,0.9999999945768);
\draw[draw=none, c0] (axis cs:5.6,0) rectangle (axis cs:6.4,0.5247374834);
\draw[draw=none, c1] (axis cs:5.6,0.5247374834) rectangle (axis cs:6.4,0.9683552465);
\draw[draw=none, c2] (axis cs:5.6,0.9683552465) rectangle (axis cs:6.4,0.971396314453);
\draw[draw=none, c3] (axis cs:5.6,0.971396314453) rectangle (axis cs:6.4,0.993346288893);
\draw[draw=none, c4] (axis cs:5.6,0.993346288893) rectangle (axis cs:6.4,0.998410311143);
\draw[draw=none, c5] (axis cs:5.6,0.998410311143) rectangle (axis cs:6.4,0.99999998783);
\draw[draw=none, c0] (axis cs:6.6,0) rectangle (axis cs:7.4,0.0015688116333);
\draw[draw=none, c1] (axis cs:6.6,0.0015688116333) rectangle (axis cs:7.4,0.0078279555453);
\draw[draw=none, c2] (axis cs:6.6,0.0078279555453) rectangle (axis cs:7.4,0.0138643829443);
\draw[draw=none, c3] (axis cs:6.6,0.0138643829443) rectangle (axis cs:7.4,0.1763508389443);
\draw[draw=none, c4] (axis cs:6.6,0.1763508389443) rectangle (axis cs:7.4,0.9938374013443);
\draw[draw=none, c5] (axis cs:6.6,0.9938374013443) rectangle (axis cs:7.4,1.0000000007393);
\end{axis}

\end{tikzpicture}%
    \caption{$N_z = 8$}
    \label{fig:varying_Nz_8}
    \end{subfigure}
    \hspace{-4em}
    \caption{Oracle skill prediction of learned latent skills $z$.
    % For $N_z \neq N_c$, where the oracle classifier predicts critical values over multiple skill features, novel behaviors are discovered as a mixture of the true skills contained in the original dataset.
    }
    \label{fig:varying_Nz}
\end{figure}
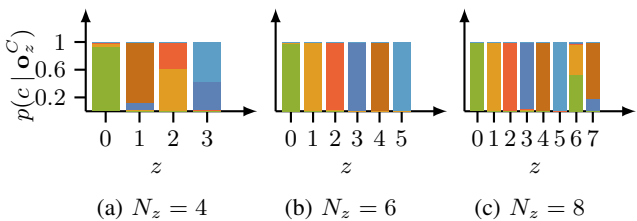

\Figref{fig:varying_Nz_6} reveals that when $N_z = N_c$, all skills in the reference dataset are extracted and assigned to a distinct latent variable $z$. 
In contrast, for $N_z \neq N_c$, it is observed that new skills emerge in a form of motion interpolation, whose existence is reflected by mixed predictions by the \oc given a latent skill $z$ (e.g. $z=2,3$ in \figref{fig:varying_Nz_4} and $z=6,7$ in \figref{fig:varying_Nz_8}).
These skills present ``mixed" features of the references.
In addition, the skill-conditioned task rewards $\rT(z)$ report consistently high value for all latent skills with different choices of $N_z$.
This indicates that the new skills also yield high task performance while presenting mixed motion patterns.
Within the scope of the defined task, these discovered skills are considered to be both \emph{novel} and \emph{useful}.

The policy training using \method can be understood as a matching procedure between the latent skills and the reference skills.
In this setting, the true skills $c \in \gC$ can be viewed as vertices that define a skill space.
The imitation reward $\rI$ motivates the policy to learn behaviors that stay within this skill space, while the skill reward $\rS$ motivates the behavior associated with each $z$ to stay far away from the averaged behavior~\citep{Zahavy2021reward}.
For $N_z = N_c$, each $z$ takes a vertex of the skill space and is assigned to a distinct $c \in \gC$.
For $N_z > N_c$, more latent variables need to be fit into the skill space.
Once the vertices are taken (pure skills), the remaining $z$ have to stay in the interior of the space (mixed skills).
For $N_z < N_c$, the latent variables have larger freedom to choose their relative position in the skill space to stay distant from each other, presenting both ``pure" and ``mixed" behaviors.

\subsubsection{Real-system Deployment}

\begin{figure}[h]
\pgfplotsset{
    width=.7\linewidth,
    height=.21\linewidth,
}

\tikzset{
    z0/.append style={draw=ourgreen, line width=2pt},
    z1/.append style={draw=ourorange, line width=2pt},
    z2/.append style={draw=ourred, line width=2pt},
    z3/.append style={draw=ourblue, line width=2pt},
    z4/.append style={draw=ourbrown, line width=2pt},
    z5/.append style={draw=ourlightblue, line width=2pt},
}
\small
    \centering
    \begin{subfigure}[]{.8\linewidth}
    % This file was created with tikzplotlib v0.10.1.
\begin{tikzpicture}

\begin{axis}[
tick align=outside,
tick pos=left,
x grid style={darkgray176},
xlabel={\(\displaystyle t\) [sec]},
xmin=-26.25, xmax=551.25,
xtick style={color=black},
y grid style={darkgray176},
ylabel={\(\displaystyle p(c \mid \mathbf{o}^{C}_{z})\)},
ymin=-0.0599999999944802, ymax=1.1,
ytick style={color=black},
clip=false,
xtick={100, 200, 300, 400, 500},
xticklabels={2, 4, 6, 8, 10},
ytick={0.2,0.6,1.0}
]
\addplot [c0]
table {plots/sgl/skill_switch/skill_switch-000.dat};
\addplot [c1]
table {plots/sgl/skill_switch/skill_switch-001.dat};
\addplot [c2]
table {plots/sgl/skill_switch/skill_switch-002.dat};
\addplot [c3]
table {plots/sgl/skill_switch/skill_switch-003.dat};
\addplot [c4]
table {plots/sgl/skill_switch/skill_switch-004.dat};
\addplot [c5]
table {plots/sgl/skill_switch/skill_switch-005.dat};

\path [z0]
(axis cs:0,1.2)
--(axis cs:75,1.2);

\path [z1]
(axis cs:75,1.2)
--(axis cs:150,1.2);

\path [z5]
(axis cs:150,1.2)
--(axis cs:225,1.2);

\path [z4]
(axis cs:225,1.2)
--(axis cs:300,1.2);

\path [z3]
(axis cs:300,1.2)
--(axis cs:375,1.2);

\path [z2]
(axis cs:375,1.2)
--(axis cs:450,1.2);

\path [z0]
(axis cs:450,1.2)
--(axis cs:525,1.2);

\node[] at (axis cs:35,1.3){$0$};
\node at (axis cs:110,1.3){$1$};
\node at (axis cs:185,1.3){$5$};
\node at (axis cs:260,1.3){$4$};
\node at (axis cs:335,1.3){$3$};
\node at (axis cs:410,1.3){$2$};
\node at (axis cs:485,1.3){$0$};

\node at (axis cs:-70,1.3){$z=$};

\end{axis}

\end{tikzpicture}%
    \end{subfigure}
    \caption{Skill sequence executed on \solo{} (top). Oracle predictions of the motions recorded from the robot (bottom).}
    \label{fig:skill_switch}
\end{figure}
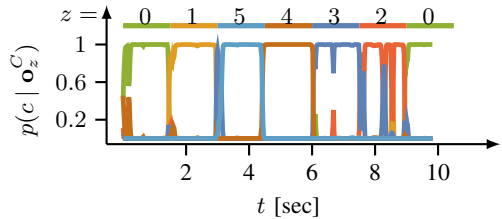

We deploy the learned policy on the real system and record 10 state transition trajectories of 120 time steps when conditioned on each skill.
The recorded motions are evaluated again with the \oc as depicted in \figref{fig:policy_deployment}.
We also illustrate the continuous change of behaviors within a trajectory by switching latent skill variables in \figref{fig:skill_switch}.
\Figref{fig:motion_sequence} provides a motion sequence on \solo{} with skill execution by the versatile policy.

\begin{figure}
    \centering
\vspace{0.7em}
\pgfplotsset{
    width=.6\linewidth,
    height=.15\linewidth,
}
    \small
    \begin{subfigure}[]{0.8\linewidth}
\tikzset{
    c0/.append style={fill=ourgreen},
    c1/.append style={fill=ourorange},
    c2/.append style={fill=ourred},
    c3/.append style={fill=ourblue},
    c4/.append style={fill=ourbrown},
    c5/.append style={fill=ourlightblue},
}
    % This file was created with tikzplotlib v0.10.1.
\begin{tikzpicture}

\begin{axis}[
tick align=outside,
tick pos=left,
x grid style={darkgray176},
xlabel={\(\displaystyle z\)},
xmin=-0.69, xmax=5.69,
xtick style={color=black},
y grid style={darkgray176},
ylabel={\(\displaystyle p(c \mid \mathbf{o}^{C}_{z})\)},
ymin=0, ymax=0.999994820473,
ytick style={color=black},
ytick={0.2,0.6,1.0},
xtick={0,1,2,3,4,5},
]
\draw[draw=none, c0] (axis cs:-0.4,0) rectangle (axis cs:0.4,0.9725112);
\draw[draw=none, c1] (axis cs:-0.4,0.9725112) rectangle (axis cs:0.4,0.97268392224);
\draw[draw=none, c2] (axis cs:-0.4,0.97268392224) rectangle (axis cs:0.4,0.972735977373);
\draw[draw=none, c3] (axis cs:-0.4,0.972735977373) rectangle (axis cs:0.4,0.974293597473);
\draw[draw=none, c4] (axis cs:-0.4,0.974293597473) rectangle (axis cs:0.4,0.999994820473);
\draw[draw=none, c5] (axis cs:-0.4,0.999994820473) rectangle (axis cs:0.4,1.000000019846);
\draw[draw=none, c0] (axis cs:0.6,0) rectangle (axis cs:1.4,0.016460488);
\draw[draw=none, c1] (axis cs:0.6,0.016460488) rectangle (axis cs:1.4,0.997625188);
\draw[draw=none, c2] (axis cs:0.6,0.997625188) rectangle (axis cs:1.4,0.9983838687);
\draw[draw=none, c3] (axis cs:0.6,0.9983838687) rectangle (axis cs:1.4,0.99876189975);
\draw[draw=none, c4] (axis cs:0.6,0.99876189975) rectangle (axis cs:1.4,0.999466505);
\draw[draw=none, c5] (axis cs:0.6,0.999466505) rectangle (axis cs:1.4,1.000000068);
\draw[draw=none, c0] (axis cs:1.6,0) rectangle (axis cs:2.4,0.00083474466);
\draw[draw=none, c1] (axis cs:1.6,0.00083474466) rectangle (axis cs:2.4,0.00487941336);
\draw[draw=none, c2] (axis cs:1.6,0.00487941336) rectangle (axis cs:2.4,0.94223071336);
\draw[draw=none, c3] (axis cs:1.6,0.94223071336) rectangle (axis cs:2.4,0.99020033736);
\draw[draw=none, c4] (axis cs:1.6,0.99020033736) rectangle (axis cs:2.4,0.99986228336);
\draw[draw=none, c5] (axis cs:1.6,0.99986228336) rectangle (axis cs:2.4,1.00000004497);
\draw[draw=none, c0] (axis cs:2.6,0) rectangle (axis cs:3.4,9.388652e-05);
\draw[draw=none, c1] (axis cs:2.6,9.388652e-05) rectangle (axis cs:3.4,0.00157572582);
\draw[draw=none, c2] (axis cs:2.6,0.00157572582) rectangle (axis cs:3.4,0.00197411622);
\draw[draw=none, c3] (axis cs:2.6,0.00197411622) rectangle (axis cs:3.4,0.99504388622);
\draw[draw=none, c4] (axis cs:2.6,0.99504388622) rectangle (axis cs:3.4,0.99535686954);
\draw[draw=none, c5] (axis cs:2.6,0.99535686954) rectangle (axis cs:3.4,1.00000002524);
\draw[draw=none, c0] (axis cs:3.6,0) rectangle (axis cs:4.4,0.00033356718);
\draw[draw=none, c1] (axis cs:3.6,0.00033356718) rectangle (axis cs:4.4,0.000372477844);
\draw[draw=none, c2] (axis cs:3.6,0.000372477844) rectangle (axis cs:4.4,0.008127271244);
\draw[draw=none, c3] (axis cs:3.6,0.008127271244) rectangle (axis cs:4.4,0.008145580434);
\draw[draw=none, c4] (axis cs:3.6,0.008145580434) rectangle (axis cs:4.4,0.999945080434);
\draw[draw=none, c5] (axis cs:3.6,0.999945080434) rectangle (axis cs:4.4,1.000000121124);
\draw[draw=none, c0] (axis cs:4.6,0) rectangle (axis cs:5.4,0.00018086094);
\draw[draw=none, c1] (axis cs:4.6,0.00018086094) rectangle (axis cs:5.4,0.00229205444);
\draw[draw=none, c2] (axis cs:4.6,0.00229205444) rectangle (axis cs:5.4,0.002299392591);
\draw[draw=none, c3] (axis cs:4.6,0.002299392591) rectangle (axis cs:5.4,0.002836998431);
\draw[draw=none, c4] (axis cs:4.6,0.002836998431) rectangle (axis cs:5.4,0.006299528131);
\draw[draw=none, c5] (axis cs:4.6,0.006299528131) rectangle (axis cs:5.4,1.000000028131);
\end{axis}

\end{tikzpicture}%
   % \end{subfigure}
%     \begin{subfigure}[]{0.49\linewidth}
% \tikzset{
%     z0/.append style={draw=ourgreen, line width=2pt},
%     z1/.append style={draw=ourorange, line width=2pt},
%     z2/.append style={draw=ourred, line width=2pt},
%     z3/.append style={draw=ourblue, line width=2pt},
%     z4/.append style={draw=ourbrown, line width=2pt},
%     z5/.append style={draw=ourlightblue, line width=2pt},
% }
%     \input{plots/sgl/skill_switch/skill_switch}
     \end{subfigure}
     
    \vspace{-.3em}
        \caption{Oracle predictions for the skills executed on the real \solo{}.}
    \label{fig:policy_deployment}
\end{figure}
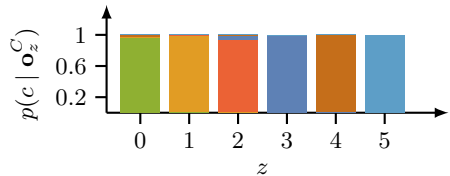

\section{Conclusion}

In this work, we propose an adversarial imitation method named \method that is capable of extracting and learning individual skills from unlabeled datasets with diverse behaviors by utilizing unsupervised skill discovery techniques.
Our results highlight the importance of the skill reward, whose absence results in skill collapses.
The experiments also indicate that \method allows extracting robust versatile policies capable of active skill control with both high fidelity and diversity with respect to the original reference motions.
Furthermore, policies can be learned with task specifications and are able to transfer to the real system without further adaptation.
Finally, our experiments confirm the emergence of novel skills that yield high task performance and present common features shared by original reference motions.

\section*{Acknowledgment}
This work is supported by the Volkswagen Stiftung (No 98 571) and the Tübingen AI Center (BMBF FKZ: 01IS18039B).

% The preferred spelling of the word ``acknowledgment'' in America is without 
% an ``e'' after the ``g''. Avoid the stilted expression ``one of us (R. B. 
% G.) thanks $\ldots$''. Instead, try ``R. B. G. thanks$\ldots$''. Put sponsor 
% acknowledgments in the unnumbered footnote on the first page.

\clearpage
\bibliographystyle{IEEEtran}
\bibliography{IEEEabrv,main}

% \clearpage
% \appendix
% \input{suppl.tex}

\end{document}